\documentclass[letterpaper]{article} % DO NOT CHANGE THIS
\usepackage{aaai2026}  % DO NOT CHANGE THIS
\usepackage{makecell}
\usepackage{times}  % DO NOT CHANGE THIS
\usepackage{helvet}  % DO NOT CHANGE THIS
\usepackage{courier}  % DO NOT CHANGE THIS
\usepackage[hyphens]{url}  % DO NOT CHANGE THIS
\usepackage{graphicx} % DO NOT CHANGE THIS
\usepackage{natbib}  % DO NOT CHANGE THIS AND DO NOT ADD ANY OPTIONS TO IT
\usepackage{caption} % DO NOT CHANGE THIS AND DO NOT ADD ANY OPTIONS TO IT
\usepackage{algorithm}
\usepackage{algorithmic}
\usepackage{booktabs}
\usepackage{longtable}

\usepackage{newfloat}
\usepackage{listings}

\DeclareCaptionStyle{ruled}{labelfont=normalfont,labelsep=colon,strut=off} % DO NOT CHANGE THIS
\floatstyle{ruled}
\newfloat{listing}{tb}{lst}{}
\floatname{listing}{Listing}
\usepackage[most]{tcolorbox}

\tcbuselibrary{listings}
\usepackage{lmodern}
\usepackage{array}
\usepackage{amsmath, amssymb}

\nocopyright
\title{LLM-as-a-Grader: Practical Insights from Large Language Model for Short-Answer and Report Evaluation}

\author{
Grace Byun\textsuperscript{\rm 1}\thanks{Equal contribution},
Swati Rajwal\textsuperscript{\rm 1}\footnotemark[1],
Jinho D. Choi\textsuperscript{\rm 1}
}

\affiliations{
\textsuperscript{\rm 1}Emory University, Atlanta, GA, USA\\
grace.byun@emory.edu, swati.rajwal@emory.edu, jinho.choi@emory.edu
}

\usepackage{bibentry}
\begin{document}

\maketitle

\begin{abstract}
Large Language Models (LLMs) are increasingly explored for educational tasks such as grading, yet their alignment with human evaluation in real classrooms remains underexamined. In this study, we investigate the feasibility of using an LLM (GPT-4o) to evaluate short-answer quizzes and project reports in an undergraduate Computational Linguistics course. We collect responses from approximately 50 students across five quizzes and receive project reports from 14 teams. LLM-generated scores are compared against human evaluations conducted independently by the course teaching assistants (TAs). Our results show that GPT-4o achieves strong correlation with human graders (up to 0.98) and exact score agreement in 55\% of quiz cases. For project reports, it also shows strong overall alignment with human grading, while exhibiting some variability in scoring technical, open-ended responses. We release all code and sample data\footnote{\url{https://github.com/emorynlp/LLM-Grading}} to support further research on LLMs in educational assessment. This work highlights both the potential and limitations of LLM-based grading systems and contributes to advancing automated grading in real-world academic settings.
\end{abstract}


\section{Introduction}
Recent advances in Large Language Models (LLMs) opened new possibilities for their application in educational contexts, including automated tutoring, feedback generation, and grading \citep{maiti2024studentsinteractllmpoweredvirtual, chiang2024largelanguagemodelassignment, chu2025enhancingllmbasedshortanswer}. Prior studies discussed how LLMs could reduce educators’ workloads by generating personalized materials and assessments, while emphasizing the importance of human–AI collaboration guided by instructors \citep{liu2025llms}. Automated grading systems, in particular, offer the potential for increased efficiency and scalability. However, their practical reliability and pedagogical value in real classrooms remains underexplored. In domains such as computational linguistics—where both factual accuracy and analytical reasoning are critical—evaluating the feasibility of LLM-based grading systems is especially important.

In this preliminary study, we investigate the use of GPT-4o \citep{openai2024gpt4ocard} to automatically evaluate short-answer quizzes and final project reports in an undergraduate Computational Linguistics course. We also release an open-source auto-grading toolkit to support reproducibility and further research. Specifically, we address the following research questions:

\begin{itemize}
    \item \textbf{RQ1}: How well do LLM-generated grades align with human evaluations?
    \item \textbf{RQ2}: What are the most common reasons for disagreement between LLM and human graders?
    \item \textbf{RQ3}: Can an open-source grading toolkit be developed to support LLM-based assessment in real-world educational settings?
\end{itemize}

To answer these questions, we collect responses from approximately 50 students across five quizzes and team project reports from a real undergraduate course. LLM's outputs are compared with grades assigned independently by two human teaching assistants (TAs). We also introduce and publicly release \textbf{LLM-as-a-Grader}, an open-source grading toolkit, along with all code and evaluation protocols, to encourage the adoption of LLM-based evaluation tools. Our findings offer insights into the strengths and limitations of using LLMs for academic evaluation and highlight considerations for their deployment in real-world classrooms.

\section{Related Work}
Prior studies have shown that LLMs can approximate human grading performance on a variety of academic tasks. LLM-generated scores correlate strongly with instructors’ grades and fall within the range of normal inter-grader variability \citep{unknown, article}. For example, ChatGPT was able to match university instructors’ exam scores within a 5–10\% margin in around 70\% of cases \cite{article}. This suggests that with proper use, LLMs can serve as reliable co-graders, reducing educators’ workload. Importantly, LLMs can also provide feedback and rationale. Several projects use AI not only to grade student answers but also to provide explanations or tutoring \citep{xie2024gradelikehumanrethinking, yeung2025zeroshotllmframeworkautomatic, miroyan2025pedagogical}. For example, one study \citep{golchin2024gradingmassiveopenonline} used chain-of-thought prompting not only to yield a score but also to generate a step-by-step reasoning for the score, effectively giving students an explanation for each point deducted. In another case, GPT-4 was used to rephrase incorrect answers from trainee teachers into correct ones, providing immediate, specific guidance on how to improve \citep{lin2024irightusinggpt}. Such findings are relevant to our study: we aim not only to assign grades but also to furnish students with formative feedback.

It is also important to mention that most prior work on LLM-based grading has relied on earlier models such as GPT-4 or GPT-3.5 \cite{chiang2024largelanguagemodelassignment, golchin2024gradingmassiveopenonline}. However, we utilize GPT-4o throughout this study due to the improved performance compared to prior models across a range of academic and reasoning benchmarks.

\begin{figure*}[htbp]
    \centering
    \setlength{\fboxrule}{0.5pt}
    \includegraphics[width=0.91\linewidth]{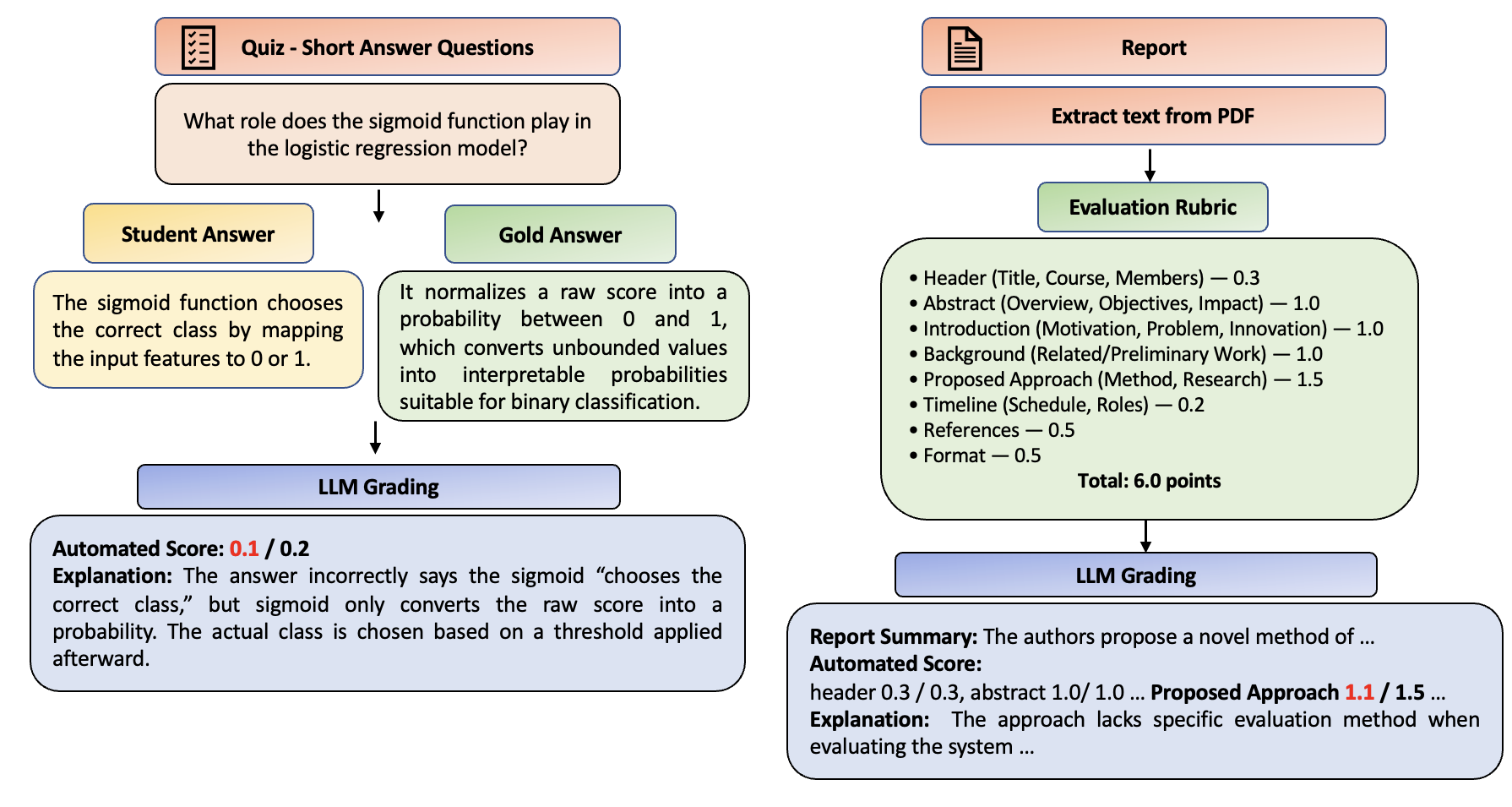}
    \vspace{-0.45em} 
    \caption{Our toolkit evaluates both short-answer quiz responses (left) and reports (right). For quizzes, a student answer is compared to a reference answer and scored based on correctness. For reports, text is extracted from PDF files and evaluated based on pre-defined rubric. Explanations are generated in both cases to justify the score.}
    \label{fig:grading_framework}
\end{figure*}

\begin{figure}[htbp]
\tiny
\begin{tcolorbox}[
    width=\columnwidth,
    colback=white,
    colframe=black,
    title=Prompt,
    sharp corners,
    listing only,
    listing options={
        basicstyle=\tiny\ttfamily,
        breaklines=true,
        breakindent=0pt,
        frame=none,
    }
]
You are grading a student's answer to a quiz question. Follow these rules:\\\\
1. The maximum score for this question is \{\textcolor{blue}{full\_score}\}.\\
2. You can only choose a score from this set: \{\textcolor{blue}{valid\_scores}\}.\\
3. Return the grade on a single line exactly in the format: “Grade: X" where X is one of \{\textcolor{blue}{valid\_scores}\}.\\
4. If you give a grade less than \{\textcolor{blue}{full\_score}\}, then on the next line provide a short single-sentence explanation starting with “Explanation:".\\
5. If you give a grade equal to \{\textcolor{blue}{full\_score}\}, do not provide any explanation or additional lines. End your response immediately after “Grade: \{\textcolor{blue}{full\_score}\}".\\
6. If the student's answer is relevant to the question and includes at least one or two key ideas from the gold answer, give the maximum score. Even if some details are missing, as long as the student shows general understanding, give the full points.\\\\
Example:\\
\textbf{Question}: All delimiters used in our implementation are punctuation marks. What types of tokens should not be split by such delimiters?\\
\textbf{Gold Answer}: URLs, email addresses, decimal numbers, and abbreviations\\
\textbf{Student Answer}: email and decimal numbers.\\
\textbf{Grade}: 0.2\\\\
7. If the student's answer is empty or obviously irrelevant, it cannot capture the essential idea, so you must give 0.\\\\
\textbf{Question}: \{\textcolor{blue}{question}\}\\
\textbf{Gold Answer}: \{\textcolor{blue}{gold\_answer}\}\\
\textbf{Student Answer}: \{\textcolor{blue}{student\_answer}\}\\
\textbf{Grade}:
\end{tcolorbox}
\caption{Prompt used to grade quiz responses. \textcolor{blue}{full\_score}\ refers to the maximum points for the question. \textcolor{blue}{valid\_scores}\ lists all possible scores the grader can assign.}
\label{fig:prompt_quiz}
\end{figure}

\section{Dataset}

\subsection{Short-answer Quiz dataset}
We collect short-answer quiz responses from around 50 undergraduates enrolled in a Computational Linguistics course. A total of five quizzes are collected over the course of the 4-month semester. Each quiz includes 10–16 open-ended questions designed to assess students' understanding of key concepts. Unlike multiple-choice questions (MCQ) format, these open-ended questions require students to articulate their understanding in their own words. For each question, a gold-standard answer is provided by the course instructor to serve as the reference for evaluation. The questions are written at the undergraduate level, covering topics such as \textit{n}-gram language models, vector space models, and basic parsing algorithms. The topics covered in each quiz along with sample questions are presented below. 
% More questions for each quiz can be found in Appendix~\ref{sec:appendix_quiz}.

Each question has a maximum score of 0.2 points, with grading done in 0.1-point increments. Since students can phrase their responses differently, grading focus on conceptual alignment with the gold answer rather than exact wording. If a student’s answer is relevant and included at least one or two key ideas from the gold answer, they receive full points, even if some details are missing. Clearly irrelevant or empty responses receive a score of zero. This grading policy requires evaluators to assess conceptual understanding rather than surface-level correctness. A graders (human or LLM) must interpret diverse expressions of the same concept and go beyond lexical matching to verify semantic alignment. This makes the task significantly more complex than evaluating fixed-format or MCQ, where correctness is binary and easier to automate.

\begin{itemize}
    \item \textbf{Quiz 1: Text Processing} \\
    Covers basic string manipulation, tokenization, normalization (lowercasing, stemming, lemmatization), and regular expressions. Example questions:
    \begin{enumerate}
        \item In which tasks can lemmatization negatively impact performance?
        \item What are the benefits and limitations of using regular expressions for tokenization vs. the rule-based tokenization approach discussed in the previous section?
    \end{enumerate}

    \item \textbf{Quiz 2: Language Models} \\
    Focuses on \textit{n}-gram language models, smoothing techniques (e.g., Laplace), and evaluation metrics such as perplexity. Example questions:
    \begin{enumerate}
        \item What indicates high entropy in a text corpus?
        \item What is the relationship between corpus entropy and language model perplexity?
    \end{enumerate}

    \item \textbf{Quiz 3: Vector Space Models} \\
    Includes term weighting (TF-IDF), cosine similarity, and document classification using vector representations. Example questions:
    \begin{enumerate}
        \item Why do we use only the training set to collect the vocabulary?
        \item What are the primary weaknesses and limitations of the K-Nearest Neighbors (KNN) classification model when applied to document classification?
    \end{enumerate}

    \item \textbf{Quiz 4: Distributional Semantics} \\
    Tests understanding of co-occurrence matrices, word-context windows, dimensionality reduction, and distributional hypothesis. Example questions:
    \begin{enumerate}
        \item What limitations does the Word2Vec model have, and how can these limitations be addressed?
        \item What are the implications of the weight matrices $W_x$ and $W_h$ in the Skip-gram model?
    \end{enumerate}

    \item \textbf{Quiz 5: Contextual Encoding} \\
    Addresses contextual word representations using pre-trained models such as BERT, including their architectural differences and contextualization mechanisms. Example questions:
    \begin{enumerate}
        \item How can one train a document-level embedding using a transformer?
        \item What are the advantages of embeddings generated by BERT compared to those generated by Word2Vec?
        \item What are the disadvantages of using BPE-based tokenization instead of rule-based tokenization? What are the potential issues with the implementation of BPE above?
    \end{enumerate}
\end{itemize}

% \begin{itemize}
%     \item \textbf{Quiz 1: Text Processing} \\
%     Covers basic string manipulation, tokenization, normalization (lowercasing, stemming, lemmatization), and regular expressions.

%     \item \textbf{Quiz 2: Language Models} \\
%     Focuses on \textit{n}-gram language models, smoothing techniques (e.g., Laplace), and evaluation metrics such as perplexity.

%     \item \textbf{Quiz 3: Vector Space Models} \\
%     Includes term weighting (TF-IDF), cosine similarity, and document classification using vector representations.

%     \item \textbf{Quiz 4: Distributional Semantics} \\
%     Tests understanding of co-occurrence matrices, word-context windows, dimensionality reduction, and distributional hypothesis.

%     \item \textbf{Quiz 5: Contextual Encoding} \\
%     Addresses contextual word representations using pre-trained models such as BERT, including their architectural differences and contextualization mechanisms.
% \end{itemize}

\subsection{Project report dataset}
In addition to quizzes, we also evaluate project reports. These are team-based assignments in which students designed and proposed an NLP project aimed at addressing a real-world problem. Each student team submit one report, resulting in a total of 14 submissions. The final deliverable is a standardized 5–8 page document that followed a consistent structure comprising sections such as abstract, motivation, problem statement, related work, technical approach, results, and conclusion. This uniform format, together with a detailed grading rubric provided by the course instructor, enables consistent extraction and evaluation of key components across all reports. The following list provides examples of selected report topics submitted by the teams.

\begin{itemize}
    \item \textbf{CAAP (Capture Assistant in Academic Papers)}: An LLM-powered tool that extracts keyphrases and definitions from academic papers, combining textual and visual information to enhance comprehension of technical content.
    
    % \item \textbf{GitFolio}: A web-based platform that converts static resumes into dynamic online portfolios using NLP, OCR, and chatbot assistance, targeting early-career professionals with limited web development experience.
    
    \item \textbf{FutureFetch}: A personalized job/internship recommendation system that extracts information from resumes using GPT-4o and filters opportunities via a custom Python backend, evaluated through both quantitative metrics and live user feedback.
    
    \item \textbf{CareerAi}: A resume enhancement and job-matching system using NLP, regular expressions, and Gemini API, designed for resume formatting and recommendation of the relevant job listings through web scraping and structured parsing.
    
    \item \textbf{MarketGuardian}: A real-time scam detection system for e-commerce listings that analyzes textual and visual cues using LLMs, reverse image search, and keyphrase extraction to flag potentially fraudulent eBay listings.
    
    \item \textbf{ATAP (Application Tracking Automation Program)}: An NLP-based tool that automates email parsing and application tracking for students, aiming to reduce inequality in opportunity access through centralized, real-time updates and support for diverse application types.
\end{itemize}

\section{Methodology}

\subsection{Prompting and scoring strategy}

\begin{table*}[ht]
\centering
\small
\renewcommand{\arraystretch}{1.1} 
\begin{tabular}{l>{\centering\arraybackslash}p{0.8cm}>{\centering\arraybackslash}p{1.3cm}>{\centering\arraybackslash}p{1.3cm}>{\centering\arraybackslash}p{1cm}>{\centering\arraybackslash}p{1.3cm}>{\centering\arraybackslash}p{1.5cm}>{\centering\arraybackslash}p{1.2cm}>{\centering\arraybackslash}p{1.5cm}}
\toprule
Quiz& $n$ & GPT Mean & Manual Mean & Mean Abs Diff & $t$-stat & $p$-value & Corr & Corr $p$-value \\
\midrule
1. Text Processing& 53 & 1.853 & 1.934 & 0.123 & -3.886 & 2.90$\times$10$^{-4}$ & 0.619 & 7.66$\times$10$^{-7}$ \\
2. Language Models& 49 & 2.076 & 2.180 & 0.104 & -9.228 & 3.28$\times$10$^{-12}$ & 0.942 & 6.20$\times$10$^{-24}$ \\
3. Vector Space Models& 52 & 1.710 & 1.738 & 0.029 & -3.638 & 6.40$\times$10$^{-4}$ & 0.967 & 1.66$\times$10$^{-31}$ \\
4. Distributional Semantics& 52 & 2.944 & 3.000 & 0.060 & -4.867 & 1.14$\times$10$^{-5}$ & 0.880 & 9.07$\times$10$^{-18}$ \\
5. Contextual Encoding& 52 & 2.694 & 2.704 & 0.029 & -1.043 & 3.02$\times$10$^{-1}$ & 0.924 & 1.47$\times$10$^{-22}$ \\
\midrule
Overall & 258 & 2.256 & 2.311 & 0.069 & -8.960 & 6.81$\times$10$^{-17}$ & \textbf{0.982} & 2.34$\times$10$^{-186}$ \\
\bottomrule
\end{tabular}
% \vspace{0.7em}
\caption{Comparison of GPT and manual grading. Statistical significance is determined using $\alpha$ = 0.05. For $t$-tests, significance indicates that difference between GPT and manual grades is not due to random chance.}
\label{tab:homework-statistics}
\end{table*}

\begin{table*}[ht!]
\centering
\small
\renewcommand{\arraystretch}{1.1} 
\begin{tabular}{l@{\hspace{1em}}c@{\hspace{1em}}c@{\hspace{1em}}c@{\hspace{1em}}c@{\hspace{1em}}c}
\toprule
\textbf{Section} & \textbf{GPT Mean} & \textbf{TA Mean} & \textbf{Mean Diff} & \textbf{Holm $p$-value} & \textbf{Significant (Holm)} \\
\midrule
Abstract      & 1.000 & 1.000 & 0.000  & N/A    & \textbf{All scores identical}  \\
Introduction  & 0.986 & 0.979 & 0.007  & 1.0000 & \textbf{False} \\
Related Work  & 0.943 & 0.943 & 0.000  & 1.0000 & \textbf{False} \\
Approach      & 1.886 & 1.986 & -0.100 & 0.0332 & \textit{True} \\
Results       & 1.750 & 1.957 & -0.207 & 0.0102 & \textit{True} \\
Conclusion    & 1.000 & 1.000 & 0.000  & N/A    & \textbf{All scores identical} \\
References    & 0.500 & 0.500 & 0.000  & N/A    & \textbf{All scores identical}  \\
Format        & 0.500 & 0.486 & 0.014  & 0.4719 & \textbf{False} \\
\bottomrule
\end{tabular}
\caption{Report Grading: Section-wise comparison of LLM and human grading using the Wilcoxon signed-rank test. Holm–Bonferroni adjusted $p$-values are shown to account for multiple comparisons. GPT-4o's section-level scores show strong alignment with those of the human grader.}
\label{tab:section_stats}
\end{table*}

\begin{table}[ht]
\centering
\renewcommand{\arraystretch}{1.1} 
\begin{tabular}{lcc}

% \begin{tabular}{l@{\hspace{1.5em}}c@{\hspace{1.5em}}c}
\toprule
\textbf{Deduction Reason Category} & \textbf{GPT(\%)} & \textbf{TA(\%)} \\ \midrule
Insufficient quantitative results & 30.8 & 15.0 \\
Superficial Related Work & 23.1 & 25.0 \\
Missing limitations discussion& 15.4 & -- \\
Formatting issue & 7.7 & 25.0 \\
Weak novelty justification & 7.7 & 10.0 \\
Lack of detail in methods & 7.7 & 5.0 \\
Weak introduction or motivation & 7.7 & -- \\
Writing quality / clarity issue & -- & 10.0 \\
Missing/inadequate conclusion & -- & 10.0 \\
\bottomrule
\end{tabular}
\caption{Comparison of categorized reasons for GPT and human deductions during report grading.}
\label{tab:deduction-comparison}
\vspace{-0.65em} 
\end{table}
 
To evaluate student quiz responses, we use a Python-based grading script that interacts with the OpenAI GPT-4o API. If a student’s response is empty, a score of 0 is assigned automatically. For all other cases, model is prompted with the quiz question, the reference answer (i.e., gold answer), and the student response. The model is instructed to assign a score from a fixed set of valid values (e.g., [0.0, 0.1, ..., 0.2]) based on how well the student’s answer matches the key ideas in the reference. If the model assigns a score less than full credit, it also provides a short explanation for the deduction. This allows us not only to generate fine-grained and transparent grading outcomes, but also to systematically analyze common error patterns and provide formative feedback that can be used to improve future iterations of the assessment.

To evaluate project reports, our autograder extracts text from PDF files using the PyMuPDF\footnote{\url{https://github.com/pymupdf/PyMuPDF}}%\cite{PyMuPDF} 
library and submits it to GPT-4o’s API for grading. Each report is assessed using a fixed rubric with eight sections: Abstract, Introduction, Related Work, Approach, Results, Conclusion, References, and Format, totaling 9 points. Each section has a specified maximum point value, and partial credit is allowed. GPT-4o is prompted to score each section individually, provide brief justifications, and generate an overall score with a short report summary.

The overall grading process is shown in Figure \ref{fig:grading_framework}. The prompts used for grading quiz is presented in Figure \ref{fig:prompt_quiz}. Two PhD students in Computer Science served as human graders. For each quiz, student responses (quiz/report) are divided between the two graders to ensure balanced evaluation. These scores are then used to compare against GPT-4o’s grading performance.

\subsection{Parameter Configuration}
For both quiz and report grading, we use the \texttt{gpt-4o} model via the OpenAI ChatCompletion API. 
To ensure deterministic and reproducible outputs, we fix \texttt{temperature} to 0.0 and \texttt{top\_p} to 1.0 in all experiments. 
For quiz grading, \texttt{max\_tokens} was set to 200 to accommodate short scoring outputs, while for report grading it was set to 1500. 
These settings minimize randomness in grading while ensuring sufficient output length for each task. 
Although fixed for consistency in our experiments, all parameters---including \texttt{temperature}, \texttt{top\_p}, and \texttt{max\_tokens}---can be freely adjusted by users to suit their specific requirements.

\section{Results}

\begin{table*}[ht!]
\small
\centering
\renewcommand{\arraystretch}{1.25}
\setlength{\tabcolsep}{6pt}
\begin{tabular}{@{}l c c | p{5.1cm} | p{5.9cm} @{}}
\toprule
\textbf{Section} & \textbf{GPT} & \textbf{TA} & \textbf{GPT Feedback} & \textbf{TA Feedback} \\
\midrule
Abstract & 1.0 & 1.0 & 
The abstract is concise and informative, clearly summarizing the project's objectives, intellectual merit, and broader societal impact. & \\
Introduction & 1.0 & 1.0 & 
The introduction clearly states the project goals and motivation, providing a well-defined problem statement. & \\
Related Work & 1.0 & 1.0 & 
Reviews relevant prior research and distinguishes the proposed work from existing solutions. & \\
Approach & 2.0 & 2.0 & 
Clearly explained and technically sound; detailed descriptions of steps and justified novelty. & \\
Results & \textbf{1.8} & \textbf{1.8} & 
Evaluation metrics and benchmarks are described properly, \textbf{but more quantitative data would help.} & 
The “Results” section should be improved. For instance, the report states that “ADAPT was evaluated using internal benchmarks,” but no details are provided about these benchmarks or \textbf{any results tables for the reader to reference. Including more detailed preliminary results would strengthen the section.} (-0.2) \\
Conclusion & 1.0 & 1.0 & 
Summarizes findings and future work effectively, reflecting on project impact. & \\
References & 0.5 & 0.5 & 
References are complete, recent, and properly formatted. & \\
Format & 0.5 & 0.5 & 
The report is well-organized and easy to read. & \\
\midrule
GPT Summary & --- & --- & 
ADAPT is an AI-powered platform designed to enhance email composition via large language models. It introduces a user-centric interface for context-aware revisions, integrating with email services for seamless communication. The project aims to improve accessibility and efficiency in digital communication, offering a novel approach to human-AI co-writing. & --- \\
\bottomrule
\end{tabular}
\caption{Section-wise Comparison of GPT and Human Grades with Feedback for Team A's Report. The grades and the reason for deduction is identical.}
\label{tab:sectionwise-t11}
\end{table*}

\begin{table*}[ht!]
\small
\centering
\renewcommand{\arraystretch}{1.25}
\setlength{\tabcolsep}{6pt}
\begin{tabular}{@{}l c c | p{5.1cm} | p{5.9cm} @{}}
\toprule
\textbf{Section} & \textbf{GPT} & \textbf{TA} & \textbf{GPT Feedback} & \textbf{TA Feedback} \\
\midrule
Abstract & 1.0 & 1.0 & The abstract is concise and informative, clearly summarizing the project's objectives, intellectual merit, and societal impact. & \\
Introduction & 1.0 & 1.0 & Clearly states the project goals and motivation with originality and well-defined context. & \\
Related Work & 1.0 & 1.0 & Reviews prior research and distinguishes the proposed work from existing solutions effectively. & \\
Approach & \textbf{1.8} & 2.0 & The proposed method is clearly explained and technically sound, with detailed descriptions of steps, models, and algorithms. \textbf{However, the novelty claim could be more explicitly justified with specific examples of how ATAP's approach differs from existing methods.} & The section is well-structured and technically solid. they could have made a stronger case for your unique contribution, but the comparisons they included already show enough distinction from prior work. Overall, it’s convincing enough to merit full marks.
\\
Results & \textbf{1.8} & \textbf{1.8} & Preliminary results are presented with a clear evaluation plan, and metrics, datasets, and benchmarks are described properly. \textbf{However, the discussion could benefit from more detailed analysis of the results and their implications.} & The results show that the system achieved an overall precision of 92.31\%, a perfect recall of 100\%, and an F1 score of 96\%. An error analysis (i.e., cases where system predictions were wrong) would have helped not only readers but also the team to improve. For instance, I see they tried to interpret the results, but including some examples (failed prediction cases) from the actual evaluation set would have been much more helpful. (-0.2)
 \\
Conclusion & 1.0 & 1.0 & Summarizes findings and reflects on impact effectively. & \\
References & 0.5 & 0.5 & References are complete, recent, and properly formatted. & \\
Format & 0.5 & 0.5 & Well-organized and easy to read, with proper formatting. & \\
\hline
GPT Summary & - & - & ATAP is an innovative application tracking tool that automates email retrieval, content classification, and status updates. It addresses inequities and improves usability for students. & - \\
\bottomrule
\end{tabular}
\caption{Section-wise comparison of GPT and human grades with feedback for Team B's project report.}
\label{tab:sectionwise-t12}
\end{table*}

\subsection{Grading comparison on quizzes}
We conduct a statistical comparison of scores assigned by LLM and a human grader across five quizzes. For each quiz, we examine the number of graded submissions, the mean scores assigned by human graders, the absolute difference between scores, paired t-test results, and Pearson correlations. Table~\ref{tab:homework-statistics} compares GPT-generated grades with manual grades. The average difference between the two sets of grades is small, ranging from 0.029 to 0.123 points. For four out of five assignments (Quiz 1 - 4), the differences are statistically significant ($p < 0.001$), meaning the observed differences are unlikely to be due to random variation, and instead reflect a systematic scoring gap between GPT and the human grader. Quiz 5 showed no significant difference ($p = 0.302$), suggesting GPT and manual grades are very similar for that assignment. Despite some mean differences, the correlation between GPT and manual grades is strong in all cases. Correlation values ranged from 0.62 to 0.97, with all values highly significant. Overall, the correlation is 0.98. This indicates that GPT grading closely tracks manual grading in ranking student performance.

To further assess the correspondence between GPT-generated scores and human evaluations, we categorize each student score based on how closely GPT matched the manual grade. Of all the cases ($n=258$), We find that GPT’s score exactly matched the human-assigned score in 55\% of cases. It assigned higher-than-human scores in 6.2\% of cases and lower-than-human scores in 38.8\%. These results suggest that while GPT-4o often aligns with human grading, it tends to be more conservative and under-grades more frequently than over-grading.

\subsubsection{Deduction Reason Analysis}
To understand evaluation behavior differences, we categorize deduction instances from both GPT-4o (13 cases) and human (20 cases) during report grading. Table~\ref{tab:deduction-comparison} shows the distribution of deduction reasons. GPT prioritizes empirical rigor, with insufficient quantitative results being the most frequent deduction (30.8\%) compared to human (15\%). Both similarly emphasize related work quality (GPT: 23.1\%, human TA: 25\%), penalizing literature reviews without critical analysis.
However, while human TA dedicates attention to formatting and presentation (25\% vs. GPT's 7.7\%) and separately assess writing quality (10\%) and conclusion adequacy (10\%), categories absent from GPT's deductions. Conversely, GPT uniquely penalizes missing limitations discussion (15.4\%).
These patterns reveal complementary evaluation approaches: GPT emphasizes analytical depth and empirical evidence, while human applies more holistic criteria including academic presentation standards. Table \ref{tab:sectionwise-t11} and \ref{tab:sectionwise-t12} present the real examples of the GPT and human evaluation. The scores and the reason for the deduction is almost the same.

Notably, GPT tends to converge with human judgment when the rubric is explicit and unambiguous (e.g., Related Work, Approach), but diverges in areas requiring stylistic or rhetorical judgment (e.g., clarity, coherence, or narrative flow). This suggests that GPT functions reliably as an effective grader for objective and content-focused criteria, consistently identifying key strengths and limitations in student work. The overlap in deduction categories with human graders further demonstrates its practical utility, indicating that GPT can serve as a dependable grading assistant. Moreover, its ability to provide structured, fine-grained feedback shows the potential for integrating GPT-based evaluation into large-scale educational settings.

\subsection{Grading comparison on project reports}
We compare section-wise scores assigned independently by GPT and a human TA to each project report. Table~\ref{tab:section_stats} reports the average scores and results of the Wilcoxon signed-rank test across the rubric sections. Across most sections, LLM’s evaluations closely match those of the human. No statistically significant differences are observed in \textit{Introduction}, \textit{Related Work}, or \textit{Format} ($p > 0.05$). In the \textit{Abstract}, \textit{Conclusion}, and \textit{References} sections, both graders assign identical scores for all submissions, suggesting high consistency in these standardized sections. However, GPT-4o assigns significantly lower scores than human evaluators in two sections: \textit{Approach} ($p = 0.0083$) and \textit{Results} ($p = 0.0020$). To control for Type I error from multiple comparisons, we apply the Holm–Bonferroni correction. After adjustment, the \textit{Approach} ($p_\text{adj} = 0.0332$) and \textit{Results} ($p_\text{adj} = 0.0102$) sections remain statistically significant. Overall, GPT-4o performs comparably to human grading, though it may be more conservative on technical or empirical sections.\\

\section{Discussion}
We develop and release an open-source grading toolkit\footnote{https://github.com/emorynlp/LLM-Grading}, 
which allows flexible configuration of model selection, number of questions, granularity, and maximum scores. 
Since the rubric is modular and the toolkit open-sourced, the framework is easily adaptable. 
Notably, grading can be performed directly from PDFs, making the system highly practical and efficient. 
These findings suggest that GPT can provide grading results that are both consistent and closely aligned with human evaluation. 

For the API cost estimate, we considered a representative case of grading \emph{Quiz 5} 
(14 questions, 50 students, totaling 700 model calls) using GPT--4o 
(Input: \$2.50/1M tokens, Output: \$10.00/1M tokens). 
With average answer and output lengths observed in our dataset, 
the total cost is well below one US dollar even without prompt caching, 
and further reduced by roughly one-third when caching the fixed rubric and question content. 
This shows that large-scale automated grading can be performed at minimal cost 
relative to the benefits in speed, consistency, and scalability. As a preliminary study, this work uses GPT-4o in the methodology. Given the promising results, we aim to extend this work in future research by incorporating additional models beyond OpenAI.

\section{Conclusion}
In this study, we explored the use of LLM to support grading in real-world classroom settings by applying GPT-4o to an undergraduate Computational Linguistics course. Our approach proved effective for both short-answer quizzes and project reports, improving grading efficiency and consistency.

GPT-4o’s grades showed strong alignment with human graders, achieving correlations up to 0.98 for quizzes and comparable section-level scores for reports. Disagreement analysis revealed that GPT-4o tended to be slightly more conservative, often under-grading relative to humans, and emphasized empirical rigor, while human graders prioritized presentation and clarity. We also developed and released \textbf{LLM-as-a-Grader}, an open-source toolkit with flexible rubrics, multiple model choices, and direct PDF grading, demonstrating its practicality in authentic classroom contexts.

These findings suggest that GPT-based grading can provide learners with objective, fine-grained feedback at scale, while reducing instructor workload and accelerating feedback turnaround. To support further research and adoption, we have released both the sample dataset and toolkit. While developed for computational linguistics, the framework is adaptable to other domains. Future work could further examine GPT’s reliability in grading multimodal submissions containing figures and equations, explore multilingual grading scenarios, and investigate the pedagogical impact of automated formative feedback on student learning outcomes.
% \section{Conclusion}
% In this study, we explored the use of LLM to support grading in real-world classroom settings by applying GPT-4o to an undergraduate Computational Linguistics course. Our approach proved effective for both short-answer quizzes and project reports, improving grading efficiency and consistency.

% Regarding our research questions, GPT-4o’s grades showed strong alignment with human graders, achieving overall correlations up to 0.98 for quizzes and comparable section-level scores for reports. Analysis of disagreement cases revealed that GPT-4o tended to be slightly more conservative, often under-grading relative to humans, and emphasized empirical rigor, while human graders focused more on presentation and clarity. Furthermore, we successfully developed and released \textbf{LLM-as-a-Grader}, an open-source toolkit that supports flexible rubrics, multiple model choices, and direct PDF grading, demonstrating its practicality for real-world educational settings.

% To support further research and adoption, we have publicly released our sample dataset and the grading toolkit. While developed for computational linguistics, the framework is easily adaptable to other educational domains, offering a practical basis for LLM-based assessment in authentic classroom contexts.

\bibliography{custom}

\end{document}